\documentclass[letterpaper, 10 pt, conference]{ieeeconf}  

\IEEEoverridecommandlockouts                              

\usepackage{mathptmx} 
\usepackage{times} 
\usepackage{amsmath} 
\usepackage{amssymb}  
\usepackage{algorithmic}
\usepackage{multicol}

\usepackage{graphicx}      
\graphicspath{ {./figures/} }

\title{\LARGE \bf
Identification of Vehicle Dynamics Parameters Using Simulation-based Inference
}

\author{Ali Boyali, Simon Thompson and David Robert Wong*
\thanks{\tt\small \{e-mails: ali.boyali, simon.thompson, david.wong\}@ tier4.jp.}}%
        
\begin{document} 

\maketitle
\thispagestyle{empty}
\pagestyle{empty}

\begin{abstract}

Identifying tire and vehicle parameters is an essential step in designing control and planning algorithms for autonomous vehicles. This paper proposes a new method: Simulation-Based Inference (SBI), a modern interpretation of Approximate Bayesian Computation methods (ABC) for parameter identification. The simulation-based inference is an emerging method in the machine learning literature and has proven to yield accurate results for many parameter sets in complex problems. We demonstrate in this paper that it can handle the identification of highly nonlinear vehicle dynamics parameters and gives accurate estimates of the parameters for the governing equations. 
\end{abstract}

\section{INTRODUCTION}

Autoware \cite{kato2015open, kato2018autoware, AutowareWeb} is the world's first open-source autonomous driving software stack. One of it's strengths is that it is largely vehicle and sensor independent, however, this means that control and calibration methods can be limited as they must be capable of working with various vehicle platforms and sensor configurations.  Accurate system identification techniques applied to specific vehicles can be costly and time-consuming. We propose a method using Sampling-Based Inference (SBI) to enable Autoware users to easily estimate vehicle parameters, enabling more performant control methods and accurate sensor-vehicle calibration. SBI can estimate parameters such as vehicle Center-Of-Gravity (COG) and tire stiffness from user captured vehicle motion data.

Vehicle COG is essential for both control and sensor-to-vehicle calibration, and tires strongly define the motion characteristics of wheeled vehicles as they interface the vehicle to the ground (road surface) and produce all the forces for the vehicle motion. 

Researchers have been putting a great deal of dedication to the theme of autonomous and robotic vehicles in the last decade. Various types of system model approximations are necessary to design planning and control algorithms.  These model approximations play a vital role in predicting the time evolution of the motion, observing the time-varying parameters and estimating the states and outputs that can not be measured directly. Estimating model parameters, especially the parameters of tires, is of enormous importance for the successful deployment of autonomous vehicles for various driving conditions.  

Vehicle models can be derived from the first-principle design or approximated by system identification methods in state-space form. The models obtained by system identification might not have a physical interpretation or connection to the motion parameters.  On the other hand, parameters that have physical interpretation are used in models derived from first principles. In this paper, we apply recent developments in machine learning to the automotive applications of parameter identification and control. 

Parameter identification is at the intersection of several scientific disciplines.  We look at the problem from the autonomous driving perspective.  The most common method used in the automotive literature is  Least Square (LS) estimation and its variants.  Other methods use filtering theory, such as the Kalman and particle filters and related Sequential Monte Carlo algorithms. Each of these methods has differing advantages and disadvantages in terms of precision and accuracy (bias-variance trade-off), algorithmic complexity and the handling of high dimensional problems. As we mainly deal with non-linearity in the state or measurements, the nonlinear least square methods are the most common choice in parameter identification. 

However, nonlinear optimization always requires careful application as the solution depends on the initial estimates and the solver capabilities. Such approaches are also prone to estimation bias (\cite{box1971bias}). The filtering methods: EKF (\cite{baffet2009estimation}), UKF (\cite{antonov2011unscented, doumiati2012vehicle}) and particle filters (\cite{bogdanski2018kalman, lundquist2013tire, berntorp2018tire}) augment the state space by endowing the parameters with random walk dynamics. The parameters are estimated together with the states. One recent method proposed for tire identification is based on the marginalization of the parameter distributions (\cite{berntorp2018tire, lundquist2013tire}). The parameter distributions are derived from the joint distribution of the states and parameters. 
 
The proposed methods in the literature are based on a simplification of either the models or noise densities.  

In parallel to those listed, in this paper, we introduce the use of a new technique for identifying vehicle parameters: Simulation-Based Inference (SBI) for parameter identification, bringing a current machine learning method for the approximate Bayesian inference into the vehicle parameter identification literature. The SBI method yields posteriors of the parameters in the form of neural density estimators (\cite{papamakarios2019neural}). The most notable contribution of the SBI method is its potential to handle many parameters simultaneously without compromising the nonlinear nature of the formulation and the form of probability density.

This feature of SBI methods fills a gap in the literature for offline parameter identification. Most of the existing methods make assumptions to simplify the identification problems, such as Gaussian density in the uncertainty, small-angle assumption, or some form of linearization. In the SBI approach, the model can assume an arbitrarily complex probability density. We show that the tire cornering stiffness parameters and the coordinates of the vehicle's center of gravity, which frequently appear in the vehicle dynamics equations, can easily be obtained by a single shot observation of the trajectory of vehicle motion. We use IMU measurements and wheel rotational speeds as input measurements. The tire parameter identification is a classical problem in the automotive literature. In addition to the tire parameters, we included the COG to increase the identification complexity and show that the SBI method can handle more parameters within a single parameter identification study. Although the COG parameters might be available for an unmodified vehicle, any modification in the mass distribution changes these parameters and might be needed to be identified. Another reason for choosing this particular set of parameters is that the lateral and longitudinal vehicle dynamics equations depend on only these parameters in the models. 

This paper especially makes the following contributions;
\begin{itemize}
    \item introducing the use of SBI methods in the parameter identification literature in automotive research, 
    \item describing the principled methodologies of the model, and parameter preparation for the identification of important parameters for the use of SBI algorithms, 
    \item providing the first experimental results of a highly nonlinear vehicle model parameter identification,
    \item providing the full uncertainty characterization of the model, possible extension and implication of these results. 
\end{itemize}

This paper is organized in the following order. We start with the details of the vehicle model in Section \ref{sec:2}. In Section \ref{sec:3}, we describe the Approximate Bayesian Computation (ABC) problem and introduce simulation-based inference and its formulation. We connect the SBI with automotive applications in Section \ref{sec:4}, specifically the estimation of tire stiffness and COG parameters from one-shot vehicle motion data. The results are given in Section \ref{sec:5} and paper is concluded in the discussion Section \ref{sec:6}.

\section{Vehicle and Tire Models } \label{sec:2}
\subsection{Vehicle Models}

Vehicle models can have different levels of complexity and degrees of freedom. Here, we refer to the vehicle motion in the lateral and longitudinal direction (vehicle body pitch, roll, and vertical motions are ignored in the fomrulation).  A full free-body diagram is given in Fig. \ref{fig:vehicle}. The subscripts \{fl, fr, rl, rr\} are used to indicate \{front-left, front-right, rear-left, rear-right\} wheel locations, and $x$ and $y$ represent the longitudinal and lateral coordinate axes respectively. One can use simplified equations by combining the front and rear tires into a single tire representation ( \cite{doumiati2012vehicle, antonov2011unscented}). Omitting the \{right and left\} subscripts by summing up the forces, we can express the resultant nonlinear equations of the vehicle motion as follows:

\begin{figure}[h]
\centering
\includegraphics[scale=0.3]{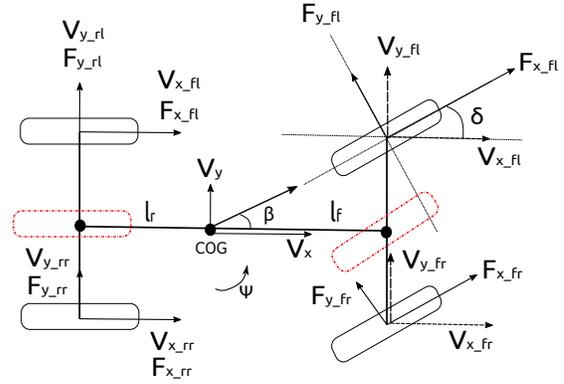}
\caption{Vehicle Bicycle Model: The body x-axis points the forward motion direction and y-axis left of the driver. The tire coordinate system is attached to each tire rolling center.}
\label{fig:vehicle}
\end{figure}

\begin{equation} \label{eq1}
\begin{split}
m(\Dot{V_x} - V_{y}\Dot{\Psi}) & = F_{xf}cos(\delta) - F_{yf}sin(\delta) + F_{xr} \\
m(\Dot{V_y} + V_{x}\Dot{\Psi}) & =F_{xf}sin(\delta) + F_{yf}cos(\delta) + F_{yr} \\
I_{z} \Ddot{\Psi}  & =l_{f}\left( F_{xf}sin(\delta) + F_{yf}cos(\delta)\right) - l_{r}F_{yr}.
\end{split}
\end{equation}
 
\noindent where $F_{x; f, r}$ and $F_{y; f, r}$ are the tire forces defined in their coordinate systems and $m$ represents the mass of the vehicle. During the simulations, we set the distance of the Center of Gravity (COG) from the rear axle center as a variable to be identified. The location of COG affects the moment of inertia $I_z$ around the $z-axis$ which is approximated by the equation $I_z = ml_{r}l_{f}$. The lateral slip $\alpha$ is a function of tire side slip angles. $\beta$ in Fig. \ref{fig:vehicle} represents the vehicle sideslip angle at the COG and $\Psi$ is the heading angle. The longitudinal slips are not consistently defined in the literature (\cite{svendenius2003review}), and we use the slip angle definitions given in \cite{pacejka2005tire} as follows:


\begin{equation} \label{eq3}
\alpha_{\{f,r\}} = -\beta_{f,r} + \delta_{f,r} = - \arctan (\frac{V_{y:f, r}}{V_{x:f, r}}) + \delta_{f,r} 
\end{equation}

\begin{equation} \label{eq4}
\kappa_{\{f, \;r\}} =\frac{R_{eff}\omega_{\{f, \;r\}} - V_{\{wx, f,r\}}}{\max(R_{eff}\omega_{\{f, \;r\}}, V_{\{wx, f,r\}})}.
\end{equation}

\noindent where $\beta_{\{f, \;r\}}$ are the wheel sideslip angles and $V_{\{x, \;y;\; f, \;r\}}$ are the tire contact patch velocities in the vehicle coordinate system. The longitudinal slips $\kappa_{\{f, \;r\}}$ are computed using the rolling speeds defined by the effective rolling radius $R_{eff}$ of the tires and the longitudinal velocities in the wheel coordinate system. The steering angle is represented by $\delta$. The tire contact patch velocities are defined following (\cite{kiencke2000automotive}):

\begin{multicols}{2}
\noindent
  \begin{equation}\label{eq5b}
   V_{yf} =  V_{y} + l_f\Dot{\Psi} 
  \end{equation}
    \begin{equation}\label{eq5c}
   V_{yr} =  V_{y} - l_r\Dot{\Psi}.
  \end{equation}
\end{multicols}

The longitudinal velocities of all tires are the same as the vehicle longitudinal velocity defined by the condition $V_{xf} = V_{xr} = V_{x}$. The effective tire radius $R_{eff}$ is computed as a linear combination of free and loaded radii $(R_f, \;R_l)$ as defined in \cite{jazar2017vehicle}:

\begin{multicols}{2}
\noindent
  \begin{equation}\label{eq6a}
   R_{eff} = \frac{2}{3}R_f + \frac{1}{3}R_l 
  \end{equation}
    \begin{equation}\label{eq6b}
   R_l  =  \frac{F_{z;\{f, r\}}}{C_{vert}}.
  \end{equation}
\end{multicols}

In Eq. \ref{eq6a}, the loaded radius $R_l$ can be measured or in the simulations, vertical tire stiffness $C_{vert}$ can be used if it is known. If the resistance forces are to be used in the simulations, such as the rolling resistance of the tires and the aerodynamic drag, the resistance forces are added to the force balance equations (\cite{boyali2006modeling}). 

\subsection{Tire Model and Dynamics} 
There are various types of tire models that are used in the literature depending on the purpose of the study. In this paper, we are interested in the determination of longitudinal and lateral tire stiffness values. The Dugoff tire model serves this purpose, parameterizing the tire forces by longitudinal and cornering stiffness (\cite{doumiati2012vehicle, rajamani2011vehicle}). The Dugoff tire model has the following form:

\begin{equation} \label{eq8}
\begin{split}
F_{\{x;f,r\}} &= C_{\{\kappa;f,r\}}\frac{\kappa_{\{f,r\}}}{1 + \kappa_{\{f,r\}}}f(\lambda).\\
F_{\{y;f,r\}} &= C_{\{\alpha;f,r\}}\frac{\tan(\alpha_{\{f,r\}})}{1 + \kappa_{\{f,r\}}}f(\lambda).\\
\end{split}
\end{equation}

\begin{equation}\label{eq9}
  f(\lambda) =
  \begin{cases}
    (2-\lambda)\lambda, & \text{if } \lambda < 1 \\
    1, & \text{if } \lambda \ge 1\\
  \end{cases}
\end{equation}

\begin{equation}\label{eq9a}
    \lambda =\frac{\mu F_{\{z;f,r\}}(1 + \kappa_{\{f, r\}})}{2\{(C_{\{\kappa;f, r\}}\kappa_{\{f, r\}})^{2} + (C_{\{\alpha;f, r\}}\tan(\alpha_{\{f, r\}}))^{2}\}^{1/2}}.
\end{equation}

\noindent where $C_{\kappa,\;\alpha}$ represent the longitudinal and cornering tire stiffness coefficients. The tire loads change due to the longitudinal acceleration and are expressed by the following equation: 
 
\begin{equation}\label{eq9b}
   F_{\{z;f,r\}} = F_{\{z\;static;f,r\}} \pm \frac{h_{cog}}{L} a_x.
\end{equation}

\noindent where the second term in the equation is the dynamic load transfer due to the acceleration and $h_{cog}$ and $L$ are the height of the COG of the vehicle from the ground and the distance between front and rear axle centers respectively.

In the tire and vehicle dynamics identification literature, the tire rolling speeds are used as an input to the equations. Here, we use tire dynamics equation and use the torque values as the inputs in the tire equations. The tire dynamics equations (\cite{doumiati2012vehicle, rajamani2011vehicle, boyali2006modeling}) are expressed as:
 
\begin{equation}\label{eq10}
    \Dot{w}_{\{f,r\}} = \left( T_{\{tr:f,r\}} + T_{\{br:f,r\}} - F_{\{x:f,r\}}R_{eff\{f,r\}} \right)  / I_{w}.
\end{equation}

\noindent where $I_w$ represents the tire rotational inertia. We represent the traction and braking torques by $T_{tr}$ and $T_{br}$ respectively. 

The longitudinal force from the output of the Dugoff model is used in this equation. The inputs are the traction $T_{tr}$ and braking torques $T_{br}$ for each of the individual tires. $I_w$  is the tire rolling moment of inertia, including the equivalent inertia of the vehicle combined with the tires (\cite{kiencke2000automotive, boyali2006modeling}). In the tire dynamics equations, delayed longitudinal forces or the delayed slip angles are used. We employed the delay dynamics on the tire slips given in the form of the equation for a better representation of the tire dynamics (\cite{heydinger1991importance, doumiati2012vehicle, rajamani2011vehicle, pacejka2005tire}):

\begin{equation}\label{eq11}
    \Dot{\hat{\alpha}}_{\{f,r\}} = \frac{V_{x;\{f,r\}}}{\sigma_l}(-\hat{\alpha}_{\{f,r\}} + \alpha_{\{f,r\}}).
\end{equation}

\noindent where we used the hat notation for the delayed dynamics. The relaxation length $\sigma_l$ is a function of the cornering $C_\alpha$ and lateral tire stiffness $K$ of the tires. It is defined in the form of $\sigma_l = C_\alpha / K$.  The relaxation length along the longitudinal axis can also be defined by analogy. However, we used the same relaxation length in the longitudinal slip delay equations. 

\subsection{Observability Conditions}
In this paper, we defined six parameters to be identified appearing in the vehicle dynamical models. These parameters are two coordinates for the COG ($h_{cog}$ and $l_f$) and four stiffness parameters for rear and front axle tires ($C_{\kappa,f,r\;|\alpha,f,r}$).  It has been shown in the literature that separately defined front and rear longitudinal stiffness values are poorly observable for the given vehicle model. As usually preferred in the literature, these values are taken to be identical and only one value for the longitudinal stiffness is identified. Our experiments also confirm this observability condition.  

In control and system identification theory, the observability concept is based on solvability conditions. These conditions yield binary decision if the parameter or internal states of the system can be solved from the given initial conditions and a set of system differential equations. 

An observability Grammian matrix is computed, and the rank conditions are checked \cite{besanccon2007nonlinear, doumiati2012vehicle}. These are the classical definitions in control theory for deterministic cases. For a system of complex nonlinear equations, when deriving the observability matrices, even if using symbolic toolboxes, it is not tractable to summarize the results \cite{doumiati2012vehicle}. In the literature, some simplifications are made on the system of differential equations. Whether the parameter or state of interest could be observable from the given set of measurements can be derived from the resulting Lie Algebraic Rank Condition (LARC). In stochastic models, if the noise in the system equations has a tractable analytical density function, the observability can be checked via the Fisher information matrix singularity condition \cite{jauffret2007observability}. 

In our solution, SBI returns a posterior density function of parameters in a neural network form. An analytical expression is not available to compute the expected value of the Fisher matrix. We verified the observability by fixing the parameter set and computing the Fisher information matrix from the expected covariance $E \left[\nabla \log \hat{P}(x|\theta^*) \nabla \log \hat{P}(x|\theta^*)^T\right]$ of the simulated score computations \cite{jauffret2007observability, alsing2019fast, alsing2018generalized}. In the computations, a fiducial parameter $\theta^*$ set is defined, and the likelihood of the simulator is approximated by the Gaussian density as it is an unknown probability function.   

In our application, we used longitudinal and lateral accelerations $(a_x, \;a_y)$ and yaw rate $\Dot{\Psi}$ as well as the tire rotational speeds ${\omega_f, \;\omega_r}$ from the vehicle's sensor outputs as the system measurements.  The sensor sampling frequency is chosen as 200 Hertz in the simulations. 

The parameter sets to be identified (given in the experiments section) for such measurements are confirmed to be observable in the statistical definition. The numerically computed Fisher Information matrix is nonsingular. 

\section{Simulation-based Inference}\label{sec:3}
\subsection{General Overview}
Given a set of observations $D = \{x_0, x_1, \ldots, x_N\}$, and  a prior density function $p(\theta)$ for the parameters $\theta$, the posterior density of the parameters $P(\theta|D)$ can be updated using Bayesian update equations if the likelihood density $P(D|\theta)$ exists. 

In the burgeoning field of likelihood-free (a.k.a simulation-based) inference methods, the problem is defined as finding the posterior $P(\theta|x=x_0)$. Suppose the likelihood function $P(x|\theta)$ is known. In that case, the posterior of the parameters can be estimated by the well-known Markov Chain Monte Carlo (MCMC) method, albeit with longer computation times if the parameter space is high-dimensional. The ABC methods deal with approximating the posterior $P(\theta|x=x_0)$ when there is no likelihood function available. The likelihood $P(x|\theta)$ is approximated from the simulated data by computing the probability of the simulated outputs $Pr(\Vert x-x_0 \Vert < \epsilon)$ around the vicinity of the observed data $x_0$. Since the chance of hitting in the probability ball defined in the high dimensional probability function is low, the likelihood-free methods usually give less accurate results when compared to the MCMC methods where the likelihood can be evaluated (\cite{papamakarios2019neural, papamakarios2016fast}). 

We consider a similar problem for vehicle parameter identification, and we aim to estimate the most likely region of the parameter space ($\theta$) that produces the observed data ($x_0$).  While doing this, we do not have an accessible likelihood function $P(x|\theta)$ except the simulator that generates observed states given the sampled parameters.  

Papamakarios et al. in \cite{papamakarios2016fast} approached the ABC problem by proposing a posterior neural density network in the form of Mixture of Density Networks (MDN) (\cite{bishop1994mixture}). In their study, they assume that the real posterior update is in the Bayesian update equation form:
 
\begin{equation}\label{eq15}
    q_{\phi}(\theta|x) \infty \frac{\Tilde{P}(\theta)}{P(\theta)} P(\theta|x).
\end{equation}
 
\noindent Here, $P(\theta)$ is a prior density whose form is known and  $\Tilde{P}(\theta)$ is the proposal density that samples the parameters for the simulator. The parameter $\phi$ represents the neural network parameters to be learnt. The authors choose analytically tractable densities such as uniform and Gaussian to evaluate the right-hand side of Eq. \ref{eq15}. The posterior density must be an MDN in this study for analytical manipulation. An algorithm is used in successive rounds for which the posterior of the previous round becomes a prior for the next round as in the active learning scheme (\cite{cranmer2020frontier}). The authors call their first simulation-based inference study SNPE\_A (Sequential Neural Posterior Estimation).  

Neural Density Estimators were proposed after the advent of the Variational Autoencoder (VAE) to model conditional densities in the literature. In VAE applications, data is fed to a neural network called an encoder. The output of the encoder is interpreted as the latent variables by which the new data points can be sampled \cite{kingma2013auto}. 
 
In the density networks, the conditional densities are modelled by applying a mask for each network layer, ensuring the autoregressive property of the conditional densities \cite{germain2015made}.  

Normalizing flows were introduced in \cite{rezende2015variational} which allow for the composition of nonlinear transformations in the neural networks. In this scheme, any arbitrarily complex probability density function can be approximated by neural networks. These are the building blocks that enabled SBI using density estimators formulated by Masked Autoregressive Flows (MAF) by  \cite{papamakarios2017masked} in which the authors introduce autoregressive networks into a normalizing flow neural network structure. 

The SNPE\_A approach was followed by a series of proposals from various authors. \cite{lueckmann2017flexible} proposed an algorithm dubbed as SNPE\_B, removing the limitation of using an MDN structure. In their algorithm, weighted prior/proposals are replaced by the weighted proportion. As a result, the posterior density can be in any form of neural density estimators.  Additionally, \cite{greenberg2019automatic} proposed SNPE\_C, which uses Automatic Posterior Transformations (APT) for posterior density estimation. Their proposition is known as the SNPE\_C in the literature. The benchmarks for these algorithms can be found in \cite{lueckmann2021benchmarking}. There have been other notable contributions in this field within the last year, as summarized by \cite{cranmer2020frontier, alsing2019fast, hermans2019likelihood, durkan2020contrastive}. 

\subsection{Automatic Posterior Transformation and SBI Toolbox}
In this study, we used SNPE\_C which directly outputs posterior density networks as it has the option of sampling the posterior parameters without using a MCMC sampler. The APT makes use of the following approximation to the posterior in the active learning scheme on top of Eq. \ref{eq15}: 

\begin{equation}\label{eq16}
    \Tilde{q}_{x,\phi}(\theta) \infty \frac{\Tilde{P}(\theta)}{P(\theta)} q_{F\phi}(\theta|x) \frac{1}{Z}
\end{equation}

\noindent where the first $q_{\phi}(\theta|x)$ term in Eq. \ref{eq15} is replaced by the proposal density term $\Tilde{q}_{x,\phi}(\theta)$. The right hand side of the real posterior $P(\theta|x)$ is replaced by its approximate density $q_{F\phi}(\theta|x)$ divided by the normalizing factor $Z$. The normalization factor is defined by:

\begin{equation}\label{eq17}
    Z = \sum q_{F\phi}(\theta|x)  \frac{\Tilde{P}(\theta)}{P(\theta)}
\end{equation}

The density $ \frac{\Tilde{P}(\theta)}{P(\theta)}\frac{1}{Z}$ on the right-hand side of Eq. \ref{eq16} is a multinomial probability density function. During the inference step, for every round of simulations and posterior updates, the simulated data sets $\{\theta_n, \;x_n\}$ are stored. A number of $K$ data pairs is chosen among them. These candidates are referred to as  K-atoms in the algorithm, and their density is assumed to be multinomial \cite{durkan2020contrastive}. 

In the active learning setting, the algorithm takes the observation $x_0$. This observation can be used to sample the parameters from the proposal density $q_{F\phi}(\theta|x=x_0)$ for the simulations along with the rounds. If the observation $x_0$ is not focused, a more general posterior is learnt, and the parameters are sampled from  $q_{F\phi}(\theta|x)$ .

\cite{tejero-cantero2020sbi} released a code repository (SBI in Python) that automatizes the parameter inference of the SNPE series and includes a neural likelihood ratio estimator, which uses neural networks to model the likelihood ratios \cite{hermans2019likelihood, durkan2020contrastive} used in MCMC problems. The code repository is still under active development. We used v0.12.1 of the SBI Toolbox, and SNPE\_C is the default method in our experiments.

\section{Parameter Identification by SBI Methods} \label{sec:4}
The SBI algorithm requires three inputs: a prior probability of the parameter of interest, a simulator, and a density network preference. The density networks are built by successive transformations in the normalizing flow structure described above. For the family of neural densities, the SBI includes \emph{nflows-Bayesian Normalizing Flows} library written for the different implementations of neural networks by \cite{durkan2019neural}. One of the recent developments in the flow-based conditional density estimator is the Neural Spline Flows (NSF) introduced in \cite{durkan2019neural}. We use this density network as the default network. 

The prior density is a multivariate uniform density for the parameters. An upper and lower bound is defined per parameter. The algorithm uses the samples drawn from this prior distribution. We used the vehicle model detailed in Section \ref{sec:2}. State and measurement trajectories for around 5 seconds with a frequency of 200 Hz are generated by this simulator. We use sufficient statistics, which is a standard method in the ABC setting, to reduce the trajectories of a point representation. This is a common practice in likelihood free inference. We used similar statistics to those the authors used in \cite{papamakarios2019sequential} for a Lotka-Volterra population model of prey and predator dynamics (i.e., log variance of each time series, auto and cross correlations). 


In our experiments, we have more than two time series obtained from the measurements. In addition, the vehicle models have control inputs and not autonomous differential equations like the Lotka-Volterra model. The input is of great importance for the identifiability of the system, and it must be designed so that the system modes are sufficiently excited. For this reason, we used a square wave with a period of five seconds for the gas and brake input and a sinusoidal input for the steering with a period of four seconds. This input profile is also used in \cite{berntorp2018tire} for tire stiffness identification.  


Alternatively, discriminating statistics can be learned by another neural network during the inference \cite{charnock2018automatic} using Information Maximizing Neural Networks (IMNN). The SBI repository  \cite{tejero-cantero2020sbi} comes with an option Embedding Network (EN), which does a similar job to obtain discriminating features from the data. In our experiments, we experimented with both methods. The use of EN yields more accurate results than the case when handcrafted sufficient statistics are used. Therefore, due to the page limitation, we give the results of the latter one in this section. 

In the experiments, we parametrize the noise by the nominal stiffness parameters $(\Delta C_{\{\alpha, \kappa; f, r\}})$ whereas the vehicle parameters are sampled from their uniform distributions. A nominal parameter set is assumed in a nominal model, which constitutes the deterministic part of our simulator. The noise or deviations from these nominal parameters (except COG parameters such as $l_f$) are used to estimate the posterior density of the noise on the parameters. Randomization is used for stochasticity. We randomize the initial longitudinal velocity of the vehicle $V_x$, which is uniformly sampled between 10-11 [m/sec], considering the real-time implementation would not exceed this range due to the safety of the experiments.  In addition, the larger this interval, the more training rounds are needed with a greater number of simulations. Another source of randomness is from the nominal parameters, which model the process noise. We generate noise for the stiffness parameters from a mixture of ten Gaussian densities with a randomly generated mean (up to 5\% of nominal values) and standard deviations (up to 5\% of nominal values). We give a set of nominal values of these parameters in Table \ref{table:tb1a}. In the experiments, at the beginning of the inference procedures, these nominal values are sampled from their prior density intervals, and are not known in advance, neither to us nor the inference algorithms. This approach shows that the whole procedure can be used for any vehicle parameter identification.

\begin{table}[h] 
\centering
\caption{Nominal Parameters}
\label{table:tb1a}
\begin{tabular}{c|llllll}
Stats/Params &$l_f$ & $h_{cog}$ & $ C_{\kappa f} $ & $C_{\kappa r}$ & $C_{\alpha f}$ & $C_{\alpha r}$\\\hline
Units & m & m&N & N & N/rad & N/rad \\\hline
Values & 1.3 & 0.5 & 1e5 & 1e5 & 6e4 & 6e4 \\
Noise Mean-Std \% & N/A & N/A & 5 & 5 & 5 & 5 \\
 
\end{tabular}
\end{table}

The inputs to the model also have similarly produced input noise. We used tire rotational speeds, vehicle body acceleration, and yaw rate in the measurements. The noise added to the measurement has a mean value of zero. The standard deviation is defined as 5\% of the measured valued for the rotational magnitudes and 10\% for the acceleration measurements. 


 

In deep learning applications, parameter normalization is a common practice. We use different sufficient statistics which may have varying magnitudes. As recommended in \cite{papamakarios2019neural} we run 1000 simulations and save the mean and standard deviations to normalize the simulator outputs for the SBI inference modules. 
 
\section{Experiments} \label{sec:5}
\begin{figure}[h]
\centering
\includegraphics[scale=0.5]{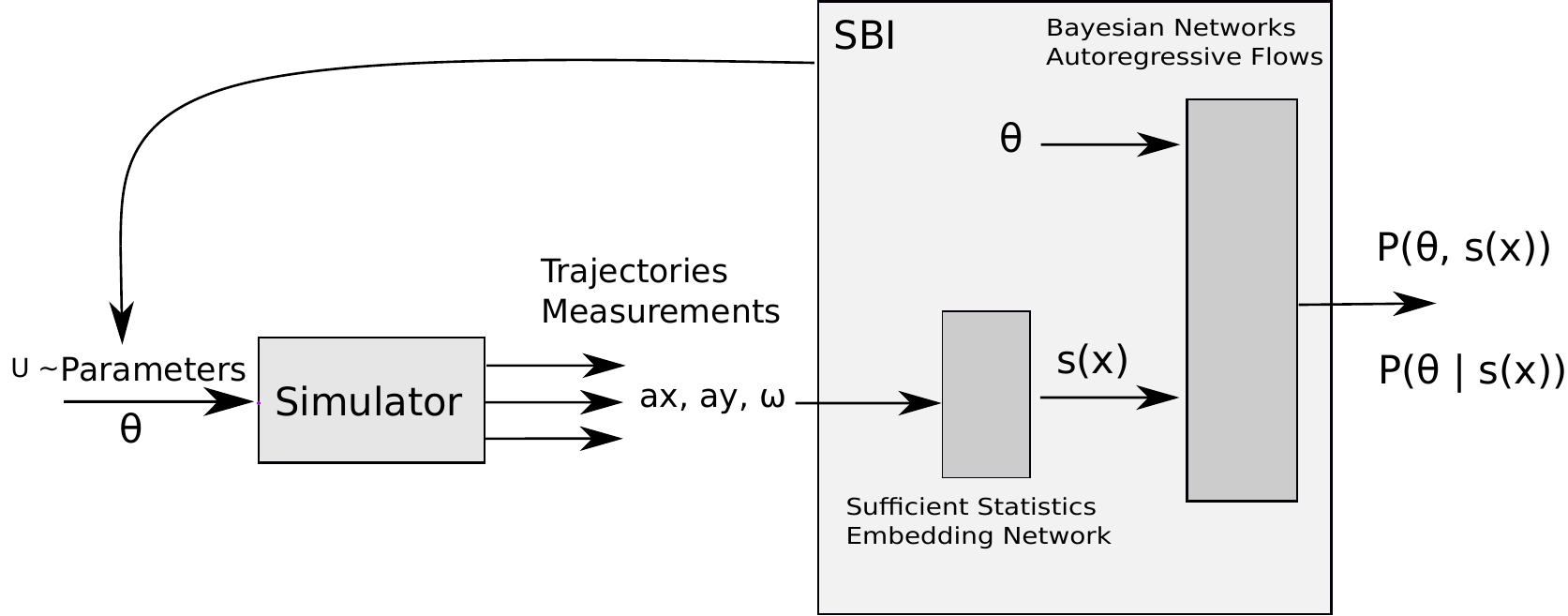}
\caption{Identification Pipeline: The prior distribution is defined by the user. The SBI module generates parameters and call the simulator. The sufficient statistics are computed and passed to the Bayesian Networks in the SBI to approximate the joint distribution and likelihood densities for the parameters of interest. }
\label{fig:pipeline}
\end{figure}

We conducted experiments with the SBI algorithm for two different studies. In the first one, using a full vehicle model including roll, pitch, and yaw body motions, we identified the lever arm for the IMU offset. In the second, we identified the vehicle COG coordinates $l_f, \;h_{cog}$ and the tire longitudinal of lateral stiffness for each axle tire set. For brevity, we present here only the results of the latter experiment.  We simulate the system and generate the observations, which are then to be used for inference. Every experiment uses 5000 simulations per round.  We use a minimum of five rounds and report the results of the posterior density yielded by the SBI algorithm. For the posterior density estimate, we sample 1000 samples from the posterior density networks. The set of measurements are the longitudinal and lateral accelerations, yaw rate, and tire rotational speeds $\{a_x, \; a_y, \; \Dot{\Psi},  \omega_f, \omega_r\}$.  We show the algorithm flow diagram in Fig. \ref{fig:pipeline}. 

\begin{figure}[h]
\centering
\includegraphics[scale=0.2]{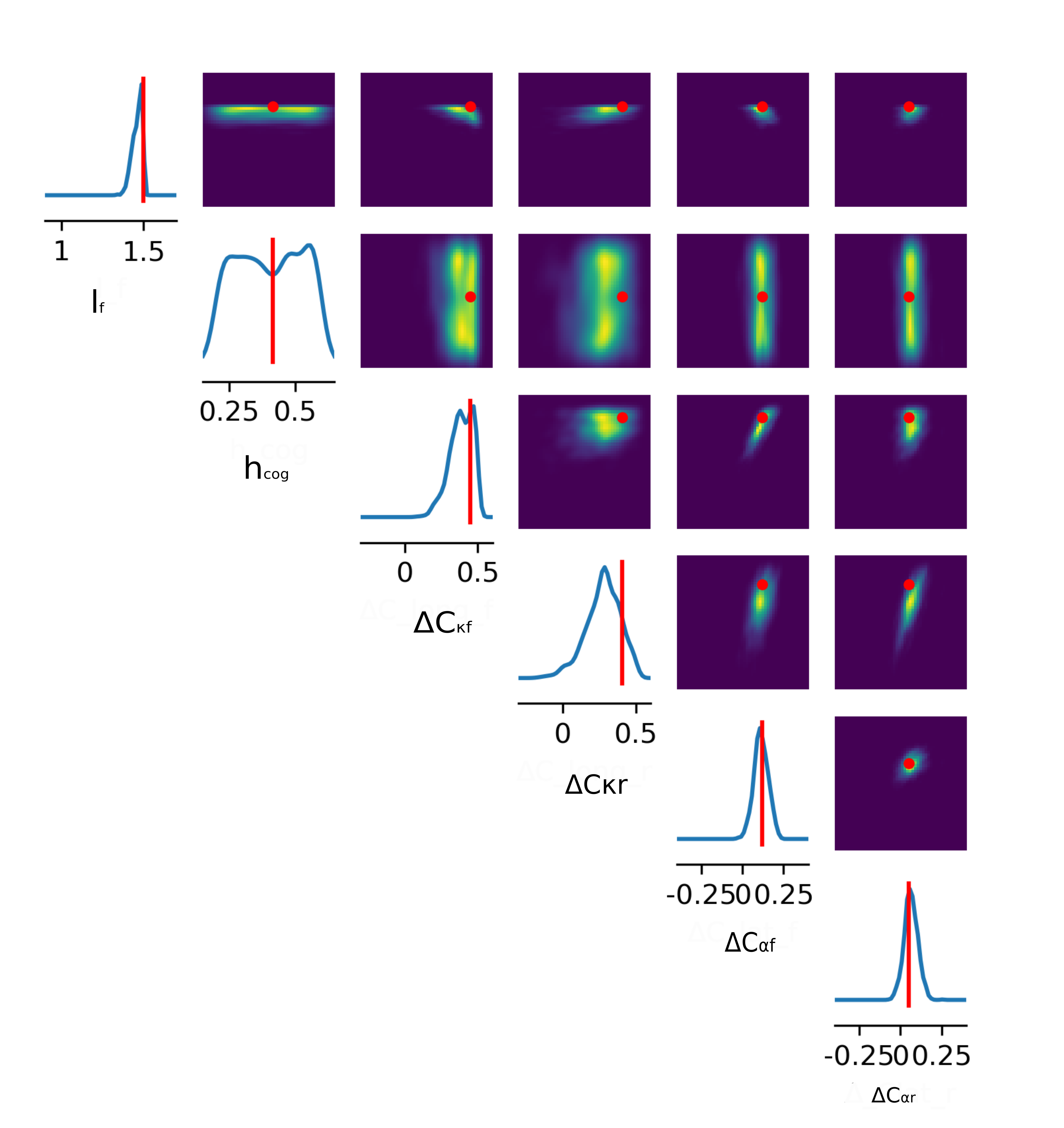}
\caption{Distribution Estimations of the Parameters: On the diagonals, the parameter posterior distributions are shown. The pair plots show the Kernel Density Estimation of the joint distributions for each pairing of parameters. The vertical red-lines mark the location of the true parameter.}
\label{fig:corner2}
\end{figure}

The posterior mean and standard deviation of the identified parameters, parameter deviations, and real values are summarised in Table ~\ref{table:tb4a}. The last row of the table shows the uniform distribution prior intervals. The corresponding pair plots are shown in Fig \ref{fig:corner2}, with the stiffness value scaled by 1e5. The pair plots show an estimated density of the parameters on the diagonal. The vertical line on each density estimate indicates the real values of the parameters. The Gaussian Kernel Density Estimates (KDE) for the pairs (joint density) are shown next to the diagonals. The red dot on the KDE graph marks the location of two parameter coordinates.

It can be seen from the Table \ref{table:tb4a} and the Fig. {\ref{fig:corner2}} that, the SBI algorithm yield accurate uncertainty estimates with a sharp peak for the distance of COG to the front axle ($l_f$) and the stiffness parameters $\Delta C_{\kappa f}$, $\Delta C_{\alpha f}$, $\Delta C_{\alpha r}$. The uncertainty of the longitudinal stiffness of the rear tires is wider. This result is compatible with the weak observability condition of the rear longitudinal stiffness described in the previous sections and our model assumptions. Our car in the simulations is front-wheel driven, and the rear tire rolls freely, exhibiting smaller traction slips than the front ones. The rear tires are only proportionally excited when braking. The height of COG parameter density has a wider distribution showing that it has bigger variation in our all experiments showing. However, even with the larger uncertainty, its mean estimate is accurate in all inference experiments. When we look at the KDE pair plots of the COG height with respect to other parameters, the location of this coordinate is restricted to be in a vertical band.   

 
\begin{table}[h] 
\centering
\caption{Parameter Priors and Posteriors}
\label{table:tb4a}
\begin{tabular}{c|llllll}
Val/Param &$l_f$ & $h_{cog}$ &$\Delta C_{\kappa f} $ & $\Delta C_{\kappa r}$ & $\Delta C_{\alpha f}$ & $\Delta C_{\alpha r}$\\\hline
Units & m & m &N & N & N/rad & N/rad \\\hline
Real Val. & 1.5     & 0.415     & 0.446     & 0.407     & 0.12 & 0.051 \\
Post. Mean & 1.46   & 0.404     &0.386      & 0.278     & 0.113 & 0.065\\
Post. Std  & 0.03   & 0.122     & 0.077     & 0.116     & 0.043 & 0.041\\
Prior Int. & 1, 1.5 &.2, .6     &-.2, .5    &-.2, .5    & $\pm$ 0.3 & $\pm$ 0.3\\
\end{tabular}
\end{table}

The identification of these parameters and their densities has two important consequences. The first one is the nominal parameter assumption before designing feedback controllers. The second importance comes from the uncertainty quantification with which robust controllers are designed. As seen from the density plots Fig \ref{fig:corner2}, the SBI gives a clear picture of the density shapes and parameters. On the other hand, the real values of the parameters are well approximated by the mean values of the densities except the rear tire longitudinal stiffness shown in Table \ref{table:tb4a}. However, the peak value of the density of this parameter almost coincides with the real value used in the simulations.

\section{CONCLUSIONS}\label{sec:6}
We proposed the use of recently emerged simulation-based inference scheme to identify vehicle and tire parameters and investigated their capabilities on a highly nonlinear vehicle simulator. The results are promising for the identification of the arbitrary number of parameters. We choose noise levels based on the parameters inspired by the literature. The measurement noise is somewhat arbitrary. In real experiments, the sensor noise model can be included in the simulator.  

The SBI method can be extended to track the parameters online, if one can use shorter time frames and different initial conditions to cover a notable portion of the state and parameter space. Density networks trained in this way might be employed for the parameter tracking application if a sufficient number of simulations are provided. This direction requires more research. 

There are well-defined methods to further increase uncertainty estimates and decrease sufficient statistics in the literature. Alsing et al. in \cite{alsing2019fast} propose an ensemble of density networks to prevent the overfitting problem. They also use a score function to reduce the number of statistics used as a feature for the trajectories. The Fisher information matrix can be obtained from the variance of the score in the simulation. It is also an important parameter for consideration in SBI studies. It can be used for the initial estimation of the parameters and the numerical evaluation of the observability conditions. More detailed explanations for the use of the Fisher information matrix in the likelihood free parameter estimation can be found in \cite{alsing2018generalized, alsing2019nuisance}. 
 
In the future, we propose to implement SBI based vehicle parameter estimation in Autoware, capture motion data from a range of vehicles, and evaluate how it improves control performance and sensor calibration across vehicle platforms, and to distribute the method in the Autoware open source project.


\section*{ACKNOWLEDGMENT}

This work was presented at the workshop Autoware – ROS-based OSS for Autonomous Driving WS26 IV2021. 

\bibliographystyle{IEEEtran}   
\bibliography{root.bib}

\end{document}